\documentclass[10pt,twocolumn,letterpaper]{article}

\usepackage[pagenumbers]{cvpr} 

\definecolor{cvprblue}{rgb}{0.21,0.49,0.74}
\usepackage[pagebackref,breaklinks,colorlinks,allcolors=cvprblue]{hyperref}
\usepackage{xcolor}
\usepackage{colortbl}
\usepackage{multirow}
\usepackage{makecell}   
\usepackage{pifont}
\usepackage{tabularray}
\usepackage{csquotes}
\usepackage{algorithm}
\usepackage{algpseudocode}
\usepackage{amsmath}
\usepackage{booktabs}

\UseTblrLibrary{diagbox}

\usepackage{amssymb}
\usepackage{utfsym}
\usepackage{wasysym}
\newcommand{\cmark}{\scalebox{1}{$\checked$}}
\newcommand{\xmark}{\scalebox{0.75}{\usym{2613}}}

\def\paperID{37228} 
\def\confName{CVPR}
\def\confYear{2026}

\title{Beyond Endpoints: Path-Centric Reasoning for Vectorized Off-Road Network Extraction}

\author{
Wenfei Guan\textsuperscript{1} \quad
Jilin Mei\textsuperscript{1} \quad
Tong Shen\textsuperscript{2} \quad
Xumin Wu\textsuperscript{2} \quad
Shuo Wang\textsuperscript{1} \quad
Chen Min\textsuperscript{1} \quad
Yu Hu\textsuperscript{1,$\dagger$}
\\[2mm]
\textsuperscript{1}Institute of Computing Technology, Chinese Academy of Sciences \\
\textsuperscript{2}Hangzhou Institute for Advanced Study, University of Chinese Academy of Sciences
\\[2mm]
{\tt\small \{guanwenfei24s, meijilin, wangshuo24z, minchen, huyu\}@ict.ac.cn} \\
{\tt\small \{shentong25, wuxumin25\}@mails.ucas.ac.cn \quad {\small $^\dagger$Corresponding author}} 
}

\begin{document}
\twocolumn[{%
    \renewcommand\twocolumn[1][]{#1}%
    \maketitle
    \vspace*{-11mm}
    \begin{center}
        \centering
        \includegraphics[width=1.0\textwidth]{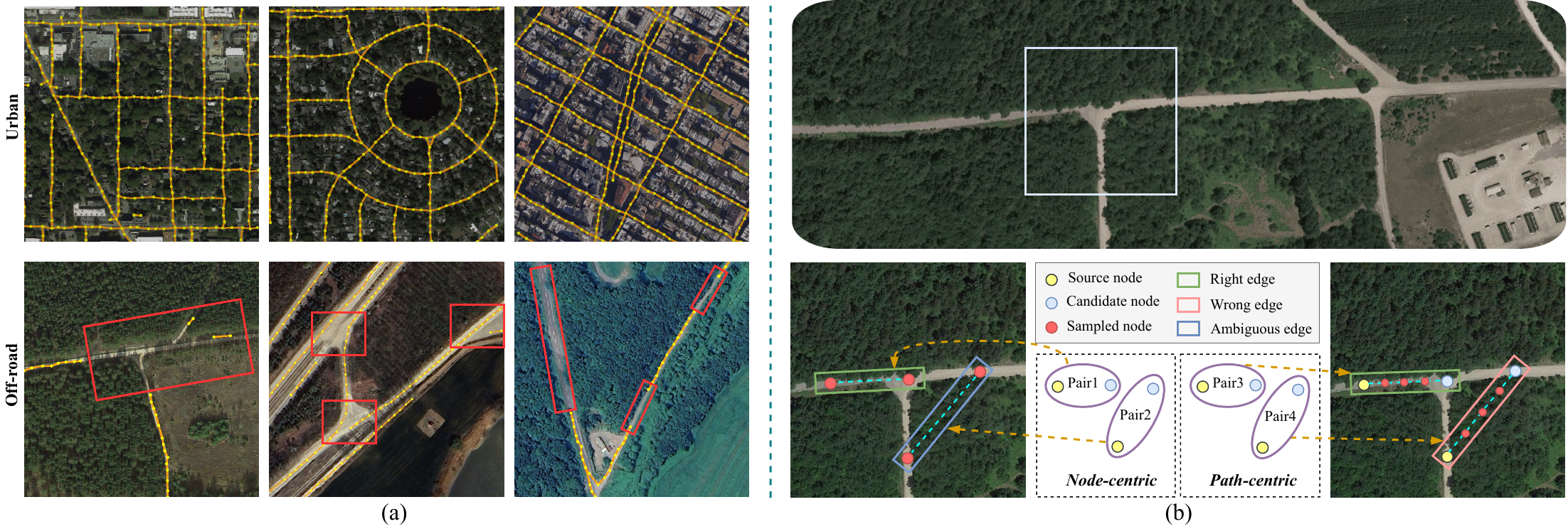}
        \captionof{figure}{
            Motivation for our work.
            \textbf{(a)} Advanced models like SAM-Road~\cite{samroad} perform well in urban environments but generate fragmented or topologically incorrect graphs in off-road scenes (failures highlighted in red), revealing a substantial domain gap and motivating our new large-scale off-road dataset.
            \textbf{(b)} Illustrating a key architectural weakness. \textit{Node-centric} models reason about features at sparse endpoints. This makes them vulnerable in ambiguous cases: while Pair1 is a clear connection, the endpoint features of Pair2 are equally plausible, yet the path itself is incorrect. In contrast, our \textit{path-centric} paradigm samples evidence along the entire path, allowing it to robustly accept correct edges (Pair3) and reject incorrect ones (Pair4), thus resolving the ambiguity. This insight motivates our path-centric approach.            }             
        \label{fig:motivation}
    \end{center}%
    \vspace{2mm}
}]
\vspace{-8pt}
\begin{abstract}

    \vspace{-0.06\baselineskip} 
    \noindent
    Deep learning has advanced vectorized road extraction in urban settings, yet off-road environments remain underexplored and challenging. A significant  domain gap causes advanced models to fail in wild terrains due to two key issues: lack of large-scale vectorized datasets and structural weakness in prevailing methods. Models such as SAM-Road \cite{samroad} employ a node-centric paradigm that reasons at sparse endpoints, making them fragile to occlusions and ambiguous junctions in off-road scenes, leading to topological errors.
    This work addresses these limitations in two complementary ways. First, we release WildRoad, a gloabal off-road road network dataset constructed efficiently with a dedicated interactive annotation tool tailored for road-network labeling. Second, we introduce MaGRoad (Mask-aware Geodesic Road network extractor), a path-centric framework that aggregates multi-scale visual evidence along candidate paths to infer connectivity robustly.
    Extensive experiments show that MaGRoad achieves state-of-the-art performance on our challenging WildRoad benchmark while generalizing well to urban datasets. An efficient vertex extraction strategy also yields roughly 2.5$\times$ faster inference, improving practical applicability. Together, the dataset and path-centric paradigm provide a stronger foundation for mapping roads in the wild.
    We release both the dataset and code at \href{https://github.com/xiaofei-guan/MaGRoad}{this repository}.
\end{abstract}    
\section{Introduction}
\label{sec:intro}

Accurate road network maps are foundational for navigation systems~\cite{mohamed2019survey}, autonomous driving~\cite{opensatmap, zhao2021tnt}, disaster response~\cite{boccardo2014remote, national2007successful}, and urban planning~\cite{rianet++}. Deep learning has enabled progress in extracting vectorized road graphs from satellite imagery, and methods such as SAM-Road~\cite{samroad} and SAM-Road++~\cite{samroadplusplus} achieve strong results on urban benchmarks like City-Scale~\cite{sat2graph}, SpaceNet~\cite{spacenet}, and Global-Scale~\cite{samroadplusplus}. Nevertheless, prior work focuses on well-structured paved roads, leaving off-road environments underexplored.

As autonomous systems move beyond cities to rural roads~\cite{ansarinejad2025autonomous}, remote sites~\cite{mining-auto}, and challenging terrains~\cite{min2024autonomous}, reliable maps become essential. Directly applying urban-trained models to off-road scenes leads to severe degradation. As shown in Fig.~\ref{fig:motivation}(a), SAM-Road performs well in urban settings but fails in off-road images, producing fragmentation, incorrect junction topology, and missed narrow or low-contrast tracks. A key reason is the absence of vectorized off-road datasets. Datasets such as DeepGlobe~\cite{deepglobe} include rural roads but provide only binary masks rather than graphs needed for topology evaluation.

Constructing a large-scale vectorized off-road dataset is crucial but prohibitively expensive~\cite{roadtracer, kelly2020maintaining}. To overcome this, we developed an efficient interactive annotation pipeline. Inspired by prompt-driven methods~\cite{sam}, our system generates initial road graph proposals from sparse user clicks on junctions and endpoints. These drafts are then refined by annotators in a web-based interface, a workflow that substantially reduces labeling time compared to fully manual annotation from scratch. Using this pipeline, we have assembled \textbf{WildRoad}, a new dataset of 221 high-resolution images (8K × 4K, 0.3 m/pixel) covering 2,100 km² across six continents.

Our WildRoad dataset brings to light the distinct challenges of off-road environments, where road segments are frequently obscured and junctions lack clear geometric structure~\cite{lourencco2023automatic, winiwarter2024extraction}. These conditions expose a key architectural weakness in prominent models like the SAM-Road series. They are built on a \textit{node-centric} paradigm, which determines connectivity by reasoning primarily about features at sparse endpoints. As visualized in Fig.~\ref{fig:motivation}(b), this reliance on endpoints is a critical flaw; The local features at candidate connection points tend to be similar and ambiguous, leaving the model with limited discriminative power to distinguish genuine road segments from spurious links without examining the entire path.

Our key insight is that robust connectivity reasoning requires a fundamental shift from node-centric to \textit{path-centric} thinking. Building on this, we propose MaGRoad, a framework designed for both robustness and efficiency. Its core module, MaGTopoNet, attains robustness by aggregating multi-scale visual evidence along the entire geodesic path of a candidate edge. This design enables the model to integrate contextual information, yielding greater resilience to weak textures and partial occlusions. Meanwhile, a unified non-maximum suppression (NMS)~\cite{neubeck2006efficient} strategy for vertex extraction ensures efficiency and reduced computational cost. The key contributions of this work are summarized below:
\begin{itemize}
\item We construct WildRoad, the first large-scale, continent-spanning vectorized benchmark for off-road environments, enabled by a novel interactive annotation pipeline that significantly reduces manual labeling effort.
\item We propose MaGTopoNet, a path-centric topology module that pools multi-scale mask evidence along edges and encodes geometric compatibility, improving graph quality on off-road and urban data.
\item We introduce an efficient vertex extraction strategy based on unified NMS that improves inference speed and scalability for large-scale mapping.
\end{itemize}   

\section{Related Work}
\label{sec:related}

\subsection{Road Network Extraction Methods}
\label{sec:road_extraction_methods}

Early deep learning approaches for road network extraction largely fell into two categories. The first, segmentation-based methods, treated the problem as pixel-wise classification~\cite{abdollahi2020vnet, chen2023semiroadexnet, gao2018end, batra2019improved, zhu2021global}. Influential models like U-Net~\cite{unet} and D-LinkNet~\cite{dlinknet} leveraged encoder-decoder architectures with features like dilated convolutions to capture multi-scale context. While achieving high pixel-level accuracy, these methods produce raster masks that lack explicit topological structure. They depend on fragile post-processing steps, such as thinning~\cite{cheng2017automatic, wen2022research, zhang1984fast}, which often introduce artifacts and fail to preserve correct road connectivity. A second category, iterative methods~\cite{rngdet, rngdet++, ventura2018iterative, yu2023iterative, vecroad}, constructs road networks in a sequential manner. Models like RoadTracer~\cite{roadtracer} and VecRoad~\cite{vecroad} start from seed points and use a learned policy to incrementally extend the graph. Although this approach can generate topologically accurate graphs, it is computationally expensive due to its auto-regressive nature and is prone to error accumulation, where an early mistake can negatively affect subsequent results.

The limitations of these earlier paradigms motivated the development of single-shot graph methods~\cite{sat2graph, shit2022relationformer, bahl2022single, xu2024patched, samroad}, which infer the entire road network in a single pass. Models like Sat2Graph~\cite{sat2graph} and TopoRoad~\cite{toporoad} pioneered techniques for encoding graph structure directly into a tensor representation for end-to-end training. Building on this, recent leading models, including SAM-Road~\cite{samroad} and SAM-Road++~\cite{samroadplusplus}, introduced a dedicated topology head called TopoNet. This module reasons about connectivity in a node-centric fashion, primarily relying on endpoint features and their geometric relationships.

\begin{table*}[h]
    \centering
    \small
    \caption{Comparison of road network extraction datasets. Bold letters denote dominant scene types (U: Urban, R: Rural, M: Mountain, W: Wild). Our WildRoad mainly focuses on wild scenes with unpaved roads, while others emphasize paved urban and suburban areas.}
    \label{tab:dataset_comparison}
    \setlength{\tabcolsep}{4pt}
    \begin{tabular}{l|c|c|c|c|c|c|c|c|c}
    \toprule
    Dataset & Label & Scene & Train & Val & Test & Size & GSD (m/p) & Area (km$^2$) & Region \\
    \midrule
    Massachusetts~\cite{mass-roads} & Raster & \textbf{U}, \textbf{R} & 1,108 & 14 & 49 & $1,500^2$ & 1.0 & 2,600 & Massachusetts \\
    DeepGlobe~\cite{deepglobe} & Raster & \textbf{U}, R & 6,226 & 243 & 1,101 & $1,024^2$ & 0.5 & 2,220 & \footnotesize Thailand, Indonesia, India  \\
    SpaceNet~\cite{spacenet} & Vector & \textbf{U} & 2,167 & -- & 567 & $400^2$ & 0.3 & 3,011 & \footnotesize Paris, Vegas, Shanghai, Khartoum \\
    City-Scale~\cite{sat2graph} & Vector & \textbf{U} & 144 & 9 & 27 & $2,048^2$ & 1.0 & 720 & \footnotesize 20 city in the U.S. \\
    Global-Scale~\cite{samroadplusplus} & Vector & \textbf{U}, \textbf{R}, M & 2,375 & 339 & 754 & $2,048^2$ & 1.0 & 13,800 & Global  \\
    \midrule
    \rowcolor{gray!15}
    WildRoad & Vector & \textbf{W}, R & 154 & 33 & 34 & $8k\times4k$ & 0.3 & 2,100 & Global  \\
    \bottomrule
    \end{tabular}
    \vspace{-3mm}
\end{table*}

While this node-centric paradigm has proven effective on structured urban benchmarks, its reliance on local endpoint information is a fundamental vulnerability. In complex environments characterized by occlusions, ambiguous boundaries, or irregular junctions, inferring connectivity from endpoints alone becomes unreliable. The failure cases shown in Fig.~\ref{fig:motivation}(a), while intensified by the urban-to-offroad domain shift, perfectly illustrate the topological errors to which this design is inherently prone, such as fragmentation and incorrect connections. This fragility motivates our work to propose a more robust, path-centric alternative. Our model, MaGRoad, features a topology head, MaGTopoNet, that explicitly aggregates visual evidence along the entire length of a potential road segment, leading to more reliable connectivity reasoning in challenging conditions.

\subsection{Datasets for Road Network Extraction}
\label{sec:datasets}

Road network datasets differ in both annotation format and geographic focus.  
\textit{Mask-annotated datasets}, such as Massachusetts Roads~\cite{mass-roads} and DeepGlobe~\cite{deepglobe}, provide pixel-level labels that are well suited for training segmentation models but lack the explicit graph structure required for topology evaluation. Massachusetts Roads covers urban, suburban, and rural regions with binary masks, while DeepGlobe offers over 8,000 satellite images at 0.5 m resolution from diverse global regions, yet its annotations remain raster-based.  
\textit{Vectorized datasets} directly support graph-based analysis and connectivity assessment. Representative examples include SpaceNet~\cite{spacenet}, which provides road vectors for several major cities worldwide, City-Scale~\cite{sat2graph}, which contains 180 high-resolution images across 20 U.S. cities, and the Global-Scale dataset~\cite{samroadplusplus}, which extends coverage to more than 3,000 regions across six continents. OpenSatMap~\cite{opensatmap} further advances this direction with lane-level annotations at 0.15 m resolution for highways and city streets. However, these resources predominantly focus on structured, well-paved urban and suburban roads. To our knowledge, no publicly available, large-scale dataset provides vectorized annotations for challenging off-road environments, a critical gap that our work aims to fill.

\subsection{Annotation Tools}
\label{sec:annotation_tools}

Traditional road mapping relies on manual digitization in GIS software such as QGIS~\cite{qgis} or ArcGIS~\cite{arcgis}, which is highly labor-intensive. Crowdsourced alternatives like OpenStreetMap~\cite{osm-weak} offer limited coverage for unpaved or off-road paths. While foundation models such as SAM~\cite{sam} have revolutionized interactive segmentation for general objects, their prompt-based capabilities have not extended to vectorized road network annotation. Existing tools lack support for direct, interactive creation and refinement of graph topology. Our work explicitly fills this gap. We introduce a novel interactive pipeline that directly integrates user prompts into the end-to-end graph generation process, enabling efficient creation and curation of our new large-scale WildRoad dataset.
\section{The WildRoad Dataset}
\label{sec:dataset}

In this section, we introduce WildRoad, a new benchmark for off-road road network extraction. The primary barrier to creating such a dataset is the prohibitive cost of manual vectorized annotation. To overcome this, we developed an AI-driven interactive pipeline and used iterative bootstrapping to efficiently curate the final collection.

\subsection{Interactive Annotation Pipeline}

\begin{figure*}[t]
    \centering
    \includegraphics[width=1.0\textwidth,keepaspectratio]{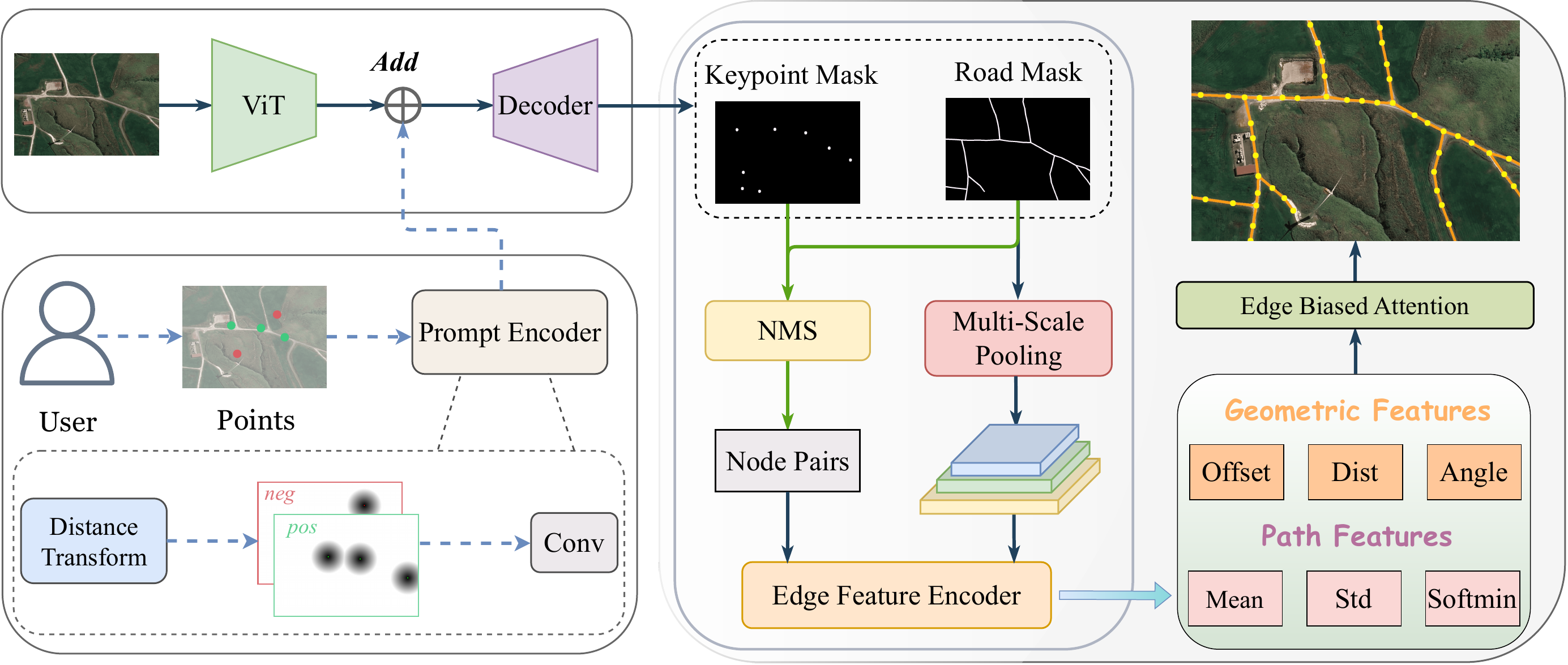}
    \caption{Overview of the MaGRoad framework. The main automated pipeline (top-left) uses a ViT encoder-decoder to produce keypoint and road probability maps. An optional interactive branch (bottom-left) encodes user clicks to guide predictions during annotation. In the path-centric graph construction module (right), candidate vertices are first extracted via NMS and paired into edges. The \textit{Edge Feature Encoder} then computes multi-scale path features by sampling the road map and combines them with geometric features. Finally, an attention mechanism processes these combined features to predict connectivity and form the final vectorized graph. \textit{Color cues:} \textcolor[HTML]{6C8EBF}{Blue} dashed lines denote the interactive branch; \textcolor[HTML]{57a018}{green} arrows indicate inference-only candidate generation steps.}    \label{fig:complete_pipeline}
\end{figure*}

Our annotation process uses a novel interactive pipeline built into a web-based interface. To accelerate labeling, the system allows annotators to provide sparse clicks at key junctions and endpoints. These clicks are processed by an Interactive Prompt Branch (blue dashed in Fig.~\ref{fig:complete_pipeline}) into spatial prompts~\cite{xu2016deep}, which guide the model's predictions. Within the tool, annotators can then refine these model-generated proposals by adding, deleting, or repositioning vertices and edges, achieving an efficient balance between automated assistance and human oversight. The system also supports high-resolution inputs and scales to large images given adequate resources. Further implementation details and visualizations of the tool are provided in the appendix.

\subsection{Dataset Bootstrapping}
Leveraging our interactive system, we constructed the off-road dataset using an efficient bootstrapping strategy with high-resolution RGB imagery sourced from Google Earth Pro~\cite{lisle2006google, mutanga2019google}. We began by annotating a small seed set of images to train an initial model. This model then generated proposals for new, unannotated regions, which human annotators corrected and refined. The newly labeled data was added to the training set to retrain and improve the model. By iterating this cycle, the model's proposals became progressively more accurate, substantially reducing the manual effort required per image. This iterative workflow enabled the efficient curation of a diverse dataset spanning forests, farmlands, deserts, and mountainous regions across six continents, including challenging cases such as tree shadows, shoreline ambiguities, and faint tracks that are difficult for existing methods to handle reliably. Tab.~\ref{tab:dataset_comparison} provides a comprehensive comparison with five representative datasets~\cite{mass-roads,deepglobe,spacenet,roadtracer,samroadplusplus}, highlighting differences in label type, road type, dominant scenes, image resolution, spatial coverage, and geographic scope.
\section{Model}
\label{sec:model}

\subsection{Overall Architecture}
\label{sec:Overview}

As illustrated in Fig.~\ref{fig:complete_pipeline}, MaGRoad's pipeline begins with a backbone encoder-decoder that produces keypoint and road-probability maps from the input image. Vertices ($V$) are extracted from the predicted masks via NMS and paired by proximity into candidate edges ($E_{\text{cand}}$). Our core module, MaGTopoNet, then filters this candidate set by scoring each edge's connectivity. It fuses geometric features with path features derived by sampling the road map, and an attention-based~\cite{vaswani2017attention, shaw2018self} classifier makes the final decision, yielding the validated edge set ($E$) that forms the road graph $G=(V,E)$. Importantly, the interactive prompt branch serves only for data annotation.

\begin{figure}[h]
    \centering
    \includegraphics[width=\columnwidth,keepaspectratio]{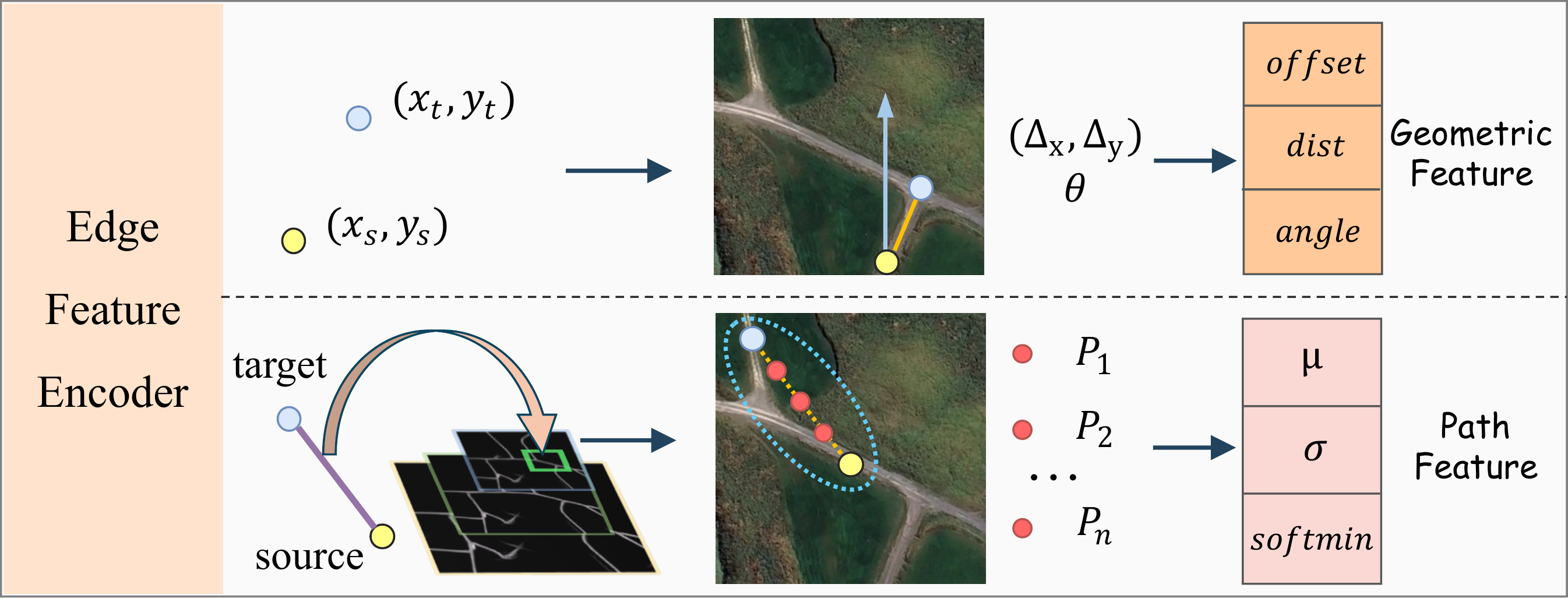}
    \caption{Edge feature encoder for connectivity prediction. 
    \textbf{Top:} Geometric features encode spatial relationships between endpoint coordinates (offset, distance, angle). 
    \textbf{Bottom:} Path features sample traversability values along the geodesic path from multi-scale road masks to compute mean, standard deviation, and softmin statistics. 
    Both are concatenated for connectivity prediction.}
    \label{fig:feature_extraction}
\end{figure}

\subsection{Edge Feature Encoding}
\label{subsec:edge}

As discussed in Sec.~\ref{sec:intro}, node-centric methods rely on local endpoint features that are often ambiguous, especially under occlusions and at irregular junctions. Our edge feature encoding addresses this by capturing evidence along the entire candidate path. For each candidate edge $(s,t)\in\mathcal{E}_{\text{cand}}$, MaGTopoNet computes a connectivity score from two complementary signals (Fig.~\ref{fig:feature_extraction}): path-based traversability from the road probability map, and geometric priors encoding the spatial relationship between endpoints.

\noindent\textbf{Geodesic path features.}
We extract path features from multi-scale road probability maps. Starting from the predicted road mask $M_{\text{road}}\in[0,1]^{H\times W}$, we generate $L=3$ scales via average pooling with kernel sizes $\{3,9,15\}$ to enhance robustness against noise and narrow occlusions. For each candidate edge $(s,t)$ connecting endpoints $s=(x_s,y_s)$ and $t=(x_t,y_t)$, we uniformly sample $N_s=32$ points $\{\mathbf{p}_i\}_{i=1}^{N_s}$ along the straight segment $\overline{st}$ (red dots in Fig.~\ref{fig:feature_extraction}) and bilinearly interpolate probability values $\{P_i^\ell\}$ from each scale $\ell$. 

At each scale, we compute three complementary statistics that capture different aspects of path quality:
\begin{equation}
\left\{\begin{aligned}
\mu_{st}^\ell &= \frac{1}{N_s}\sum_{i=1}^{N_s} P_i^\ell, \\
\sigma_{st}^\ell &= \sqrt{\frac{1}{N_s}\sum_{i=1}^{N_s} (P_i^\ell - \mu_{st}^\ell)^2}, \\
\operatorname{softmin}_{st}^\ell &= -\frac{1}{\tau}\log\sum_{i=1}^{N_s}\exp\bigl(-\tau(1-P_i^\ell)\bigr),
\end{aligned}\right.
\label{eq:path_stats}
\end{equation}
where $\mu_{st}^\ell$ measures average traversability, $\sigma_{st}^\ell$ quantifies along-path consistency (low values indicate uniform road likelihood), and $\operatorname{softmin}_{st}^\ell$ with temperature $\tau=5.0$ emphasizes potential bottlenecks by penalizing low-probability regions. Concatenating these statistics across all scales yields the path feature $\mathbf{f}_{\text{path}}^{st}\in\mathbb{R}^{3L}$.

\noindent\textbf{Geometric features.}
In addition to path cues, we encode spatial properties of each candidate. For $(s,t)$ with $(x_s,y_s)$ and $(x_t,y_t)$, we compute: (i) offsets $\Delta x,\Delta y$ normalized to $[-1,1]$; (ii) Euclidean distance $d_{st}$; and (iii) bearing angle $\theta_{st}=\arctan2(\Delta y,\Delta x)$ encoded with Fourier features $\{\sin(m\theta),\cos(m\theta)\}_{m=1}^4$. These yield $\mathbf{f}_{\text{geo}}^{st}\in\mathbb{R}^{11}$.

\subsection{Edge-Biased Attention}
\label{subsec:attention}

Given the encoded edge features, the remaining challenge is to select the correct connections from each source vertex's candidate set. We address this with a self-attention mechanism that introduces geometric competition among candidates. Concretely, we form an edge token by concatenating $\mathbf{f}_{\text{geo}}^{st}$ and $\mathbf{f}_{\text{path}}^{st}$ and project to a hidden dimension $D_h=256$. Within each per-source candidate set, self-attention~\cite{shaw2018self} uses an additive bias matrix $B$ to inject geometric priors:
\begin{equation}
\mathbf{A} = \operatorname{softmax}\!\left(\frac{\mathbf{Q}\mathbf{K}^\top}{\sqrt{d}} + \mathbf{B}\right),
\end{equation}
where $B_{ij}=-\lambda_{\text{comp}}\,\mathbb{I}[i\neq j]$ applies a uniform negative bias to all off-diagonal pairs. This competition term encourages sparse edge selection by penalizing simultaneous activation, allowing the model to select true connections while suppressing spurious pairings. The refined tokens are mapped to connectivity logits via an MLP head.

\subsection{Efficient Vertex Extraction}

We optimize the pipeline that converts predicted masks into candidate vertices. Previous approaches~\cite{samroad,samroadplusplus} apply NMS independently to keypoint and road masks and then merge results with an additional NMS pass, requiring three separate suppression stages. We unify this into a single NMS pass by concatenating candidates from both masks, with a score offset ($+0.9$) for keypoint candidates to ensure their priority during suppression.

Beyond this structural simplification, we address the core computational bottleneck in the NMS inner loop. The standard implementation performs batch array operations per neighbor, including fancy indexing, temporary array allocation, and batch memory writes, all of which incur substantial overhead despite $O(Nk\log N)$ overall complexity. Our refactored algorithm first sorts all candidates by score and then, for each point, directly suppresses every lower-scoring neighbor via scalar operations, avoiding intermediate arrays entirely. Together, these modifications achieve a 2.5$\times$ speedup on WildRoad (Tab.~\ref{tab:wild_results}), offering a trade-off that boosts the F1 score when network completeness is prioritized over topological precision.

\section{Experiments}
\label{sec:experiments}

\subsection{Datasets}
\label{subsec:datasets}

We evaluate MaGRoad on four benchmarks: our WildRoad for off-road scenarios, and three established urban datasets: \textit{City-Scale}~\cite{sat2graph}, \textit{SpaceNet}~\cite{spacenet}, and \textit{Global-Scale}~\cite{samroadplusplus}. WildRoad emphasizes challenging off-road terrains with heavy occlusion and faint tracks, while urban benchmarks assess generalization to structured road networks. Dataset statistics are provided in Tab.~\ref{tab:dataset_comparison}.

\subsection{Evaluation Metrics}
We adopt graph-topology metrics that assess connectivity and geometric accuracy. APLS (Average Path Length Similarity)~\cite{spacenet} measures graph similarity by comparing optimal path lengths between sampled point pairs, with values in $[0,1]$ where 1 indicates perfect agreement. TOPO~\cite{biagioni2012inferring} evaluates topology correctness by matching vertices within a distance threshold and measuring precision and recall of both vertices and edges. These metrics complement each other: APLS emphasizes path validity for navigation, while TOPO measures local graph structure fidelity.

\begin{table*}[t]
  \centering
  \caption{Quantitative results on the WildRoad benchmark. MaGRoad sets a strong baseline. MaGRoad-\textit{fast} denotes the version with our efficient vertex extraction strategy, which achieves the highest F1 score and a 2.5$\times$ speedup, introducing a trade-off with APLS.}
  \label{tab:wild_results}
  \setlength{\tabcolsep}{6pt}
  \begin{tabular}{l|ccccc|c}
    \toprule
    Method & P$\uparrow$ & R$\uparrow$ & F1$\uparrow$ & APLS$\uparrow$ & APLS+F1$\uparrow$ & Time (min)$\downarrow$ \\
    \midrule
    Sat2Graph~\cite{sat2graph} & 83.92 & 57.50 & 68.11 & 48.73 & 116.84 & 133.1 \\
    SAM-Road~\cite{samroad} & 87.20 & 68.65 & 76.61 & 68.71 & 145.32 & \underline{73.3} \\
    SAM-Road++~\cite{samroadplusplus} & 87.52 & 68.69 & 76.74 & \underline{69.72} & 146.46 & 76.1 \\
    \midrule
    MaGRoad & \underline{88.45} & \underline{71.48} & \underline{78.85} & \textbf{72.56} & \underline{151.41} & 74.9 \\
    MaGRoad-\textit{fast} & \textbf{90.93} & \textbf{75.43} & \textbf{82.22} & 69.29 & \textbf{151.51} & \textbf{27.8} \\
    \bottomrule
  \end{tabular}
  \vspace{-3mm}
\end{table*}

\subsection{Implementation Details.}
\label{subsec:implementation}

We implement MaGRoad using a ViT-B~\cite{dosovitskiy2020image} backbone pretrained on SAM~\cite{sam}. Training uses patch-based sampling: 1024×1024 patches with batch size 4 for our WildRoad dataset to capture sparse road structures, 512×512 patches with batch size 16 for City-Scale and Global-Scale, and 256×256 patches with batch size 64 for SpaceNet. We sample 256 source vertices per patch for WildRoad and 512 for urban datasets to balance computational cost with topology coverage.
For WildRoad, the segmentation branch is supervised by a combined Dice~\cite{jadon2020survey} and weighted BCE loss (positive weight 10 to handle class imbalance), while the topology head uses standard BCE loss. We employ the Adam optimizer with an initial learning rate of 1e-3 for randomly initialized components and 1e-4 for the pretrained ViT encoder, decaying by 0.1 at 80\% of total training epochs. Standard augmentations include random 90-degree rotations and spatial cropping.

Key hyperparameters are domain-specific: for WildRoad, we use a candidate search radius of $r{=}200$ pixels and multi-scale path pooling kernels of $\{3, 9, 15\}$ to handle occlusions and sparse networks; for urban scenes, these are set to $r{=}64$ and $\{1, 5, 9\}$, respectively. The geodesic path sampling uniformly interpolates $N_s{=}32$ points along each candidate edge. At inference, mask and topology classification thresholds are tuned to maximize F1 score on the validation set. All experiments were conducted on four NVIDIA RTX 6000 GPUs.

\subsection{Results on WildRoad}
We evaluate MaGRoad against leading methods on our challenging WildRoad test set. The quantitative results, presented in \cref{tab:wild_results}, show that our baseline method establishes a new state-of-the-art. Notably, it surpasses the previous best, SAM-Road++, in both graph completeness (F1 score) and topological accuracy (APLS), underscoring the effectiveness of its core design. Furthermore, our optimized version, MaGRoad-\textit{fast}, pushes the F1 score even higher, achieving 82.22 through an efficient vertex extraction strategy, which we analyze in detail in Sec.~\ref{sec:faster_extraction}.

\begin{figure*}[t]
  \centering
  \includegraphics[width=\linewidth]{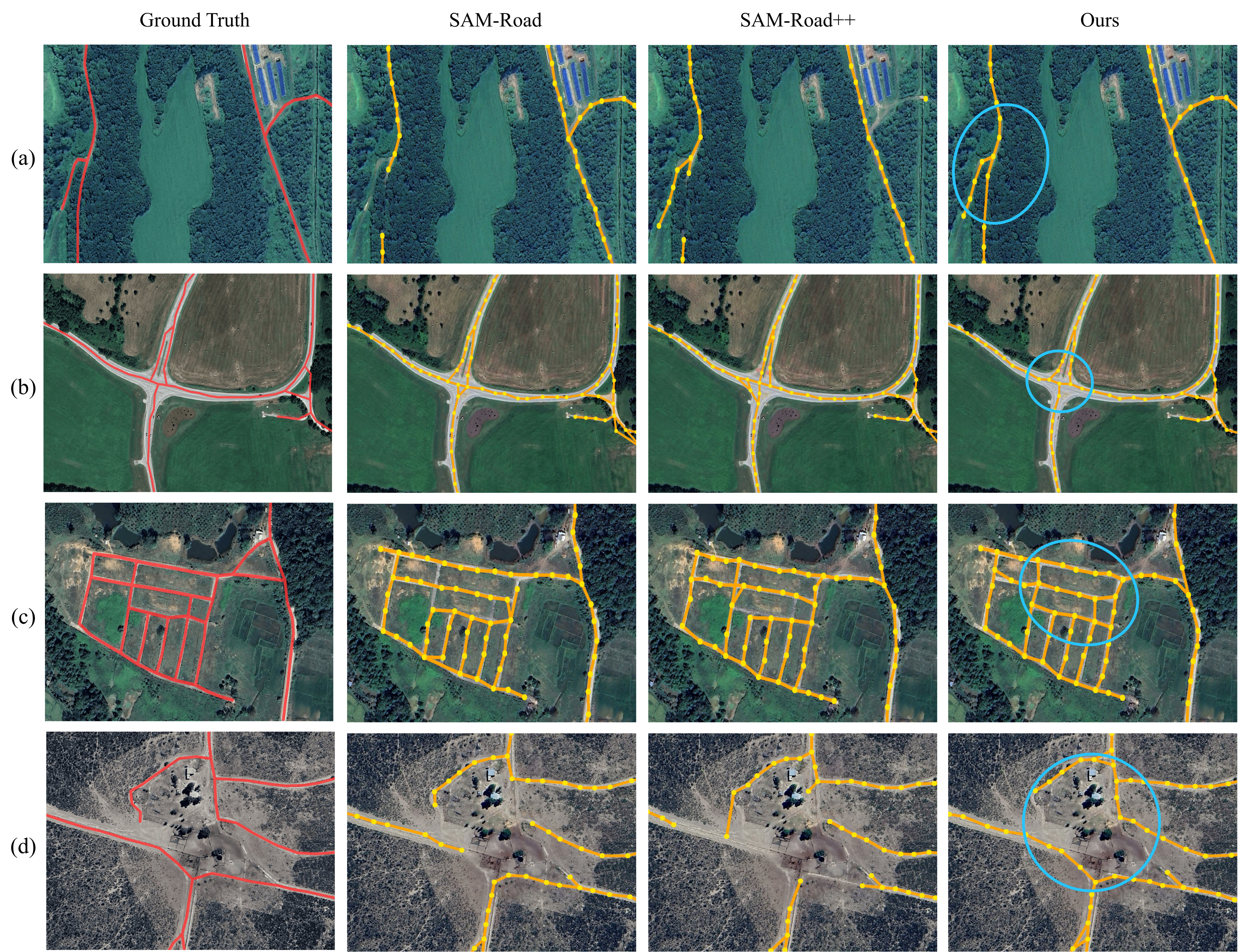} 
  \caption{Visual comparison of road network predictions on challenging scenes. Please zoom in and view in color. Our method exhibits superior robustness across diverse scenarios. In (a) and (b), it yields more accurate topology around complex crossroads and curved intersections. In (c), it achieves more complete connectivity in low-contrast residential areas, while in (d), it better preserves continuity in unstructured environments. Blue circles indicate regions where our method outperforms previous approaches.}
  \label{fig:ours_comparison}
\end{figure*}

The visual comparisons in \cref{fig:ours_comparison} provide direct insight into this quantitative advantage. Node-centric baselines like SAM-Road and SAM-Road++ consistently produce fragmented graphs where roads are occluded by tree cover (\cref{fig:ours_comparison}a) and fail to resolve the correct topology at complex, non-standard junctions (\cref{fig:ours_comparison}b). In contrast, MaGRoad's ability to aggregate evidence along the entire path allows it to maintain connectivity through these occlusions and correctly infer the network's structure in sparsely connected residential areas (\cref{fig:ours_comparison}c) and unstructured dirt tracks (\cref{fig:ours_comparison}d). These results confirm that our path-centric approach is more robust to the visual ambiguity in off-road scenes, directly translating to superior performance.

\subsection{Generalization to Urban Datasets}
To confirm that our path-centric design is a generalizable improvement and not overfit to off-road scenes, we evaluated MaGRoad on three urban datasets. As shown in \cref{tab:urban_datasets}, the results reveal a consistent and insightful trend.

\begin{table}[h]
  \centering
  \small
  \caption{Quantitative comparison on City-Scale, SpaceNet, and Global-Scale datasets. MaGRoad achieves competitive performance across all datasets.}
  \label{tab:urban_datasets}
  \setlength{\tabcolsep}{5pt}
  \renewcommand{\arraystretch}{1.15}
  \begin{tabular}{l|l|cccc}
      \toprule
      & Method & P$\uparrow$ & R$\uparrow$ & F1$\uparrow$ & APLS$\uparrow$ \\
      \midrule
      \multirow{5}{*}{\rotatebox{90}{\parbox{2.2cm}{\centering \textit{City-Scale}}}} 
          & Sat2Graph~\cite{sat2graph}   & 80.70 & 72.28 & 76.26 & 63.14 \\
          & RNGDet++~\cite{rngdet++}     & 85.65 & 72.58 & \underline{78.44} & 67.76 \\
          & SAM-Road~\cite{samroad}      & \textbf{90.47} & 67.69 & 77.23 & \underline{68.37} \\
          & SAM-Road++~\cite{samroadplusplus}
                                        & \underline{88.39} & \textbf{73.39} & \textbf{80.01} & 68.34 \\
          & \cellcolor{gray!15}MaGRoad   
          & \cellcolor{gray!15}84.46 
          & \cellcolor{gray!15}\underline{72.66} 
          & \cellcolor{gray!15}78.11 
          & \cellcolor{gray!15}\textbf{71.27} \\
      \midrule
      \multirow{5}{*}{\rotatebox{90}{\parbox{2.2cm}{\centering \textit{SpaceNet}}}}
          & Sat2Graph~\cite{sat2graph}   & 85.93 & \underline{76.55} & 80.97 & 64.43 \\
          & RNGDet++~\cite{rngdet++}     & 91.34 & 75.24 & \underline{82.51} & 67.73 \\
          & SAM-Road~\cite{samroad}      & \underline{93.03} & 70.97 & 80.52 & 71.64 \\
          & SAM-Road++~\cite{samroadplusplus}
                                        & \textbf{93.68} & 72.23 & 81.57 & \textbf{73.44} \\
          & \cellcolor{gray!15}MaGRoad   
          & \cellcolor{gray!15}87.72 
          & \cellcolor{gray!15}\textbf{81.01} 
          & \cellcolor{gray!15}\textbf{84.23} 
          & \cellcolor{gray!15}\underline{72.29} \\
      \midrule
      \multirow{5}{*}{\rotatebox{90}{\parbox{2.2cm}{\centering \textit{Global-Scale} \\ (In-Domain)}}} 
          & Sat2Graph~\cite{sat2graph} & \underline{90.15} & 22.13 & 35.53 & 26.77 \\
          & RNGDet++~\cite{rngdet++} & 79.02 & 45.23 & 55.04 & 52.72 \\
          & SAM-Road~\cite{samroad} & \textbf{91.93} & 45.64 & 59.80 & 59.08 \\
          & SAM-Road++~\cite{samroadplusplus} & 88.95 & \underline{49.27} & \underline{62.33} & \textbf{62.19} \\
          & \cellcolor{gray!15}MaGRoad
          & \cellcolor{gray!15}80.01 
          & \cellcolor{gray!15}\textbf{52.90} 
          & \cellcolor{gray!15}\textbf{62.68} 
          & \cellcolor{gray!15}\underline{60.16} \\
      \midrule
      \multirow{5}{*}{\rotatebox{90}{\parbox{2.2cm}{\centering \textit{Global-Scale} \\ (Out-of-Domain)}}}
          & Sat2Graph~\cite{sat2graph} & \textbf{84.73} & 19.75 & 30.64 & 22.49 \\
          & RNGDet++~\cite{rngdet++} & 70.22 & 35.71 & 47.34 & 38.08 \\
          & SAM-Road~\cite{samroad} & \underline{84.54} & 33.81 & 46.64 & 40.51 \\
          & SAM-Road++~\cite{samroadplusplus} & 82.21 & \underline{36.04} & \textbf{48.34} & \textbf{43.17} \\
          & \cellcolor{gray!15}MaGRoad
          & \cellcolor{gray!15}77.27 
          & \cellcolor{gray!15}\textbf{36.47} 
          & \cellcolor{gray!15}\underline{48.23} 
          & \cellcolor{gray!15}\underline{41.26} \\
      \bottomrule
  \end{tabular}
  \vspace{-3mm}
\end{table}

Across all urban datasets, MaGRoad demonstrates a distinct and consistent strength in recall, achieving the highest scores on SpaceNet and both splits of Global-Scale. This tendency towards producing more complete road networks is a direct benefit of our path-centric paradigm, which excels at identifying true connections by aggregating rich visual evidence along an entire path rather than relying solely on ambiguous endpoint features.

This high recall translates into highly competitive or state-of-the-art F1 scores. On SpaceNet, for instance, MaGRoad achieves the best overall F1 score. While precision-focused models like SAM-Road++ show their own strengths, our method provides a powerful alternative for applications where network completeness is paramount. These results confirm that path-centric reasoning is a fundamentally robust and valuable approach for road network extraction across diverse environments.

\subsection{Ablation Studies} 
We conduct ablation studies on the WildRoad test set to systematically evaluate the contribution of each key component in our model. The results are presented in \cref{tab:ablation_components}.

\noindent\textbf{Analysis of MaGTopoNet Components.}
Our primary ablation examines the synergy among path features (\textit{P}), geometric features (\textit{G}), and edge-biased attention (\textit{E}) in our full path-centric model (Exp 6). Removing any component causes a significant performance drop. Most critically, omitting path features (Exp 4) drops APLS by nearly 10 points, confirming that aggregating evidence along the path is essential for topology reasoning. Geometric features and edge-biased attention provide powerful complementary priors; removing them also degrades performance, with attention proving especially vital for encoding geometric compatibility and achieving a high F1 score.

\noindent\textbf{Path-centric vs. Node-centric Features.}
To understand the distinct roles of these feature types, we compare our path-centric model (Exp 6) against a strong node-centric baseline (Exp 7), following the approach of SAM-Road~\cite{samroad}. As shown in the table, our path-centric approach outperforms the node-centric baseline on both F1 and, most critically, the topological metric APLS. This demonstrates that for challenging off-road scenes, reasoning along the entire path is superior for ensuring both topological integrity and correct local connectivity.

\begin{table}[t]
  \centering
  \small
  \caption{Ablation study on the WildRoad test set, analyzing the   contributions of different feature types. \textit{N}: Node feature, \textit{P}: Path feature, \textit{G}: Geometric feature, \textit{E}: Edge-biased attention. Exp 6 represents our full path-centric model.}
  \label{tab:ablation_components}
  \setlength{\tabcolsep}{5pt}
  \begin{tabular}{l|cccc|cccc}
    \toprule
    Exp & \textit{N} & \textit{P} & \textit{G} & \textit{E} & P$\uparrow$ & R$\uparrow$ & F1$\uparrow$ & APLS$\uparrow$ \\
    \midrule
    1 & \xmark & \cmark & \xmark & \xmark & 79.83 & 69.74 & 74.27 & 53.90 \\
    2 & \xmark & \cmark & \xmark & \cmark & 84.62 & 63.30 & 75.97 & 63.77 \\
    3 & \xmark & \xmark & \cmark & \xmark & 80.09 & 62.01 & 70.36 & 62.66 \\
    4 & \xmark & \xmark & \cmark & \cmark & 82.45 & 62.96 & 71.24 & 63.07 \\
    5 & \xmark & \cmark & \cmark & \xmark & 86.77 & 66.94 & 75.36 & 68.10 \\
    \rowcolor{gray!12}
    6 & \xmark & \cmark & \cmark & \cmark & 88.45 & 71.48 & 78.85 & 72.56 \\
    \midrule
    \rowcolor{gray!12}
    7 & \cmark & \xmark & \cmark & \cmark & 88.16 & 69.48 & 77.53 & 69.51 \\
    \rowcolor{gray!12}
    8 & \cmark & \cmark & \cmark & \cmark & 87.86 & 69.20 & 77.23 & 69.07 \\
    \bottomrule
  \end{tabular}
\end{table}

\begin{figure}[t]
  \centering
  \includegraphics[width=\linewidth]{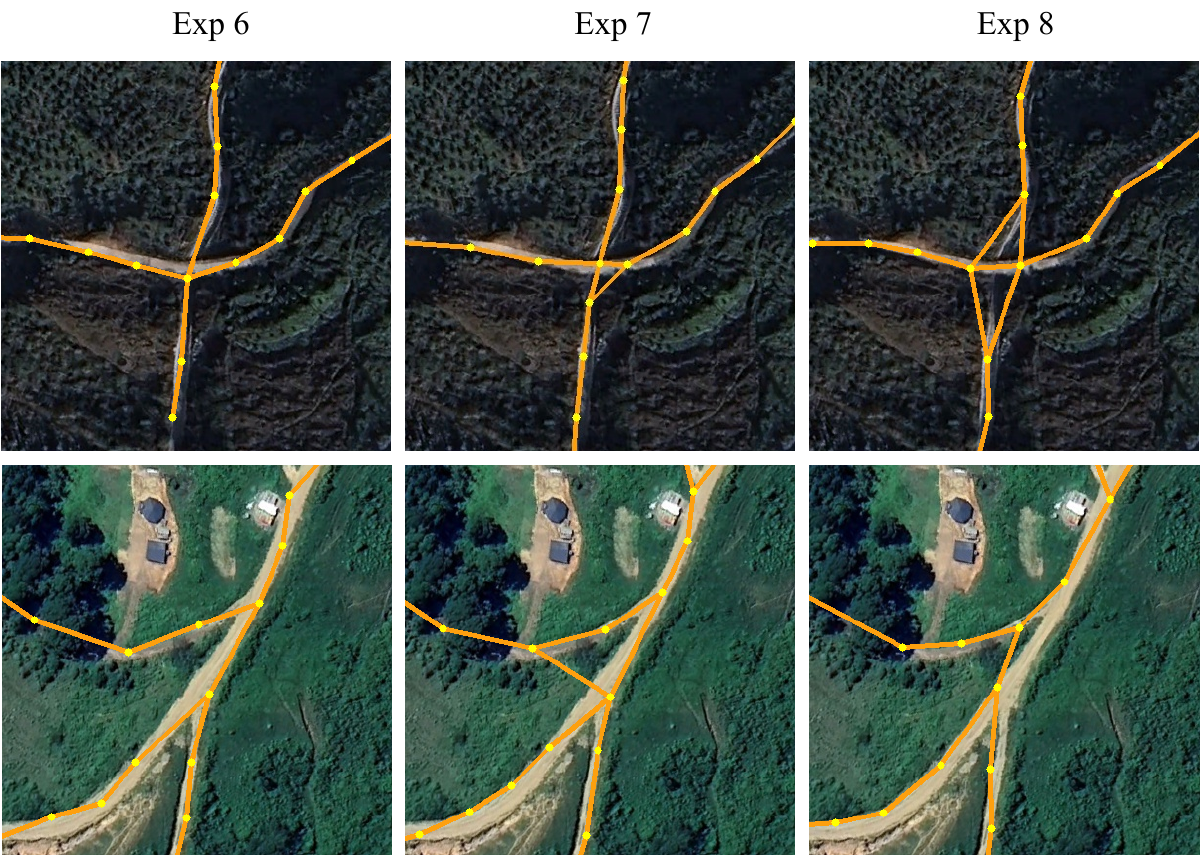}
  \caption{Visual comparison of key ablation models. The node-centric model (Exp 7) produces erroneous topological \enquote{shortcuts}, whereas our path-centric model (Exp 6) correctly infers the underlying structure, demonstrating its robustness to ambiguity.}
  \label{fig:ablation_vis}
\end{figure}

Interestingly, combining both feature types (Exp 8) does not improve results. As visualized in \cref{fig:ablation_vis}, the combined model inherits the failure modes of the node-centric approach, producing similar topological errors to Exp 7. This suggests that for visually ambiguous scenes, the explicit signal from our path features provides a more robust basis for classification than the implicit information from endpoints, reinforcing that an explicit path-centric paradigm is a more effective approach for this task.

\begin{table}[h]
  \centering
  \caption{Ablation study on multi-avg pool configurations.}
  \label{tab:ablation_multiavgpool}
  \resizebox{\columnwidth}{!}{
  \begin{tblr}{
    column{even} = {c},
    column{3} = {c},
    column{5} = {c},
    row{2-3} = {c}, 
    vline{1} = {-}{},
    vline{2} = {-}{},
    vline{7} = {-}{},
    hline{1,4} = {-}{0.08em},
    hline{2} = {-}{0.05em},
  }
  \diagbox{metrics}{kernel} & \{1,7,13\} & \{3,9,15\} & \{1,5,9\} & \{3,9\} & \{9\} \\
  APLS               & 70.43          & 72.56          & 70.35         & 68.88       & 69.17    \\
  F1                 & 77.26          & 78.85          & 78.61         & 75.99       & 75.97     
  \end{tblr}
  }
\end{table}

\noindent\textbf{Multi-Scale Path Aggregation.}
We also study the configuration of the multi-scale average pooling for path feature extraction. As shown in \cref{tab:ablation_multiavgpool}, using pooling kernels of sizes3, 9, 15 yields the best performance. This configuration effectively balances fine-grained local details and broader context. Other multi-scale settings like {1, 7, 13} and {1, 5, 9} result in a noticeable performance drop, particularly in APLS. More importantly, using fewer scales such as {3, 9} or single-scale {9} leads to a significant degradation in both metrics. These findings validate our multi-scale design, demonstrating it is crucial for handling variable road widths and occlusions.

\noindent\textbf{Efficient Vertex Extraction.}
\label{sec:faster_extraction}
Finally, we analyze our efficient vertex extraction strategy, which replaces the multi-stage NMS with a single, unified pass. As shown in \cref{tab:wild_results}, this optimized model (MaGRoad-\textit{fast}) demonstrates a key trade-off between vertex recall and topological precision. The unified NMS is less aggressive, yielding a denser vertex set. This significantly boosts recall and pushes the F1 score to a new state-of-the-art of 82.22, at the cost of minor topological noise that lowers the APLS score. For applications where throughput and network completeness are critical, MaGRoad-\textit{fast} offers a compelling 2.5$\times$ speedup and a higher F1 score, highlighting our framework's flexibility.

\section{Conclusion} 
\label{sec:conclusion}

We addressed off-road vectorized road extraction by introducing WildRoad, the first continent-spanning benchmark, and MaGRoad, a path-centric framework aggregating multi-scale evidence along candidate paths for robust connectivity inference. MaGRoad achieves state-of-the-art results on WildRoad and generalizes to urban datasets, demonstrating that path-centric reasoning is a stronger foundation for road extraction beyond urban settings.
\section*{Acknowledgments}
This work was supported by National Natural Science Foundation of China under Grant No.U23B2034 and No.62176250, Beijing Natural Science Foundation (L259015 and L259049), and the Innovation Program of Institute of Computing Technology, Chinese Academy of Sciences under Grant No. 2024000112.

{
    \small
    \bibliographystyle{ieeenat_fullname}
    \bibliography{main}
}

\clearpage
\setcounter{page}{1}
\maketitlesupplementary

\section{Training the Interactive Model}
\label{sec:suppl_interactive_system}

A core component of our annotation pipeline is the interactive model that generates graph proposals from sparse user clicks. To achieve this, we employ an end-to-end training paradigm that learns to interpret user intent without requiring real-time human interaction during the training phase. The primary challenge is bridging the gap between a static training setup and a dynamic interactive task. We overcome this by introducing a strategy to simulate user prompts, enabling the model to learn the mapping from sparse spatial cues to dense road network structures.

\subsection{Simulated Prompt Generation}
During training, we simulate user clicks by automatically sampling positive and negative points from the ground-truth annotations. 
\begin{itemize}
    \item \textbf{Positive prompts} are sampled from topologically significant locations on the ground-truth graph. Specifically, we identify all vertices that are either junctions (degree $>$ 2) or endpoints (degree $=$ 1). A random subset of these keypoints is selected to serve as positive guidance for the model.
    \item \textbf{Negative prompts} are sampled from background regions to prevent spurious graph generation. To ensure these points are not sampled too close to road boundaries, which could create ambiguity for the model, we first establish a buffer zone by dilating the ground-truth road mask with a radius of $\text{dist}_{\min}$. Negative points are then exclusively sampled from the area outside this buffer, representing unambiguous non-road regions.
\end{itemize}
To enhance the model's robustness to imprecise clicks, we apply a minor random spatial jitter to the coordinates of all sampled points. This process is detailed in Algorithm~\ref{alg:prompt_simulation}.

\subsection{Prompt Encoding and Feature Fusion}
The simulated point prompts are transformed into a spatial representation that can be fused with the image features. Given a set of prompt coordinates $P = \{p_i\}$, we first compute a distance transform map $D \in \mathbb{R}^{H \times W}$, where each pixel $(u,v)$ stores the minimum Euclidean distance to any point in $P$. This distance map is then processed by a shallow convolutional encoder (two 3$\times$3 conv layers) to produce a multi-channel prompt feature map, $F_{\text{prompt}}$. This feature map is fused with the image features, $F_{\text{image}}$, from the main ViT encoder via element-wise addition. The resulting fused features, $F_{\text{fused}} = F_{\text{image}} + F_{\text{prompt}}$, are then passed to the geometry decoder, effectively guiding the final road graph prediction as shown in the interactive branch of our main pipeline (see Fig.~2 in the main paper).

\subsection{Implementation Details}
For reproducibility, we specify the key hyperparameters for the prompt simulation. We sample up to $N_{pos}=10$ positive prompts, with the number of negative prompts set by a 1:1 ratio. The buffer radius for negative sampling is $\text{dist}_{\min} = 50$ pixels, and a spatial jitter with a standard deviation of 3.0 pixels is applied to all prompt coordinates. The interactive training is conducted on 1024$\times$1024 patches, and critical training parameters such as batch size, learning rate, and optimizer settings are kept identical to those of the baseline automated model. The only additions are the prompt simulation and feature fusion steps; the loss function and optimization process remain unchanged, highlighting the efficiency of our approach.

\begin{algorithm}[t!]
    \caption{Positive and Negative Prompt Generation}
    \label{alg:prompt_simulation}
    \begin{algorithmic}[1]
        \Require 
            GT graph $G=(V, E)$; A $\text{dist}_{\text{min}}$ radius;
            Number of positive prompts $N_{pos}$ and negative prompts $N_{neg}$.
        \vspace{4pt}
        \Ensure 
            A set of simulated and jittered prompt points $P$.
        \Statex
        \Function{SimulatePromptPoints}{$G, N_{pos}, N_{neg}$}
            \State $V_{\text{key}} \gets \{v \in V \mid \text{degree}(v) \neq 2\}$
            \State $P_{\text{pos}} \gets \text{RandomSample}(V_{\text{key}}, N_{pos})$
            \State
            \State $M_{\text{road}} \gets \text{RasterizeToMask}(G)$
            \State $K \gets \text{CreateStructuringElement}(\text{radius}=\text{dist}_{\text{min}})$
            \State $M_{\text{buffer}} \gets \text{MorphologicalDilation}(M_{\text{road}}, K)$
            \State $A_{\text{background}} \gets \neg M_{\text{buffer}}$
            \State $P_{\text{neg}} \gets \text{RandomSampleFromArea}(A_{\text{background}}, N_{neg})$
            \State
            \State $P_{\text{combined}} \gets P_{\text{pos}} \cup P_{\text{neg}}$
            \State $P \gets \text{ApplySpatialJitter}(P_{\text{combined}})$
            \State \Return{$P$}
        \EndFunction
    \end{algorithmic}
\end{algorithm}

\begin{figure*}[t]
    \centering
    \includegraphics[width=1.0\linewidth]{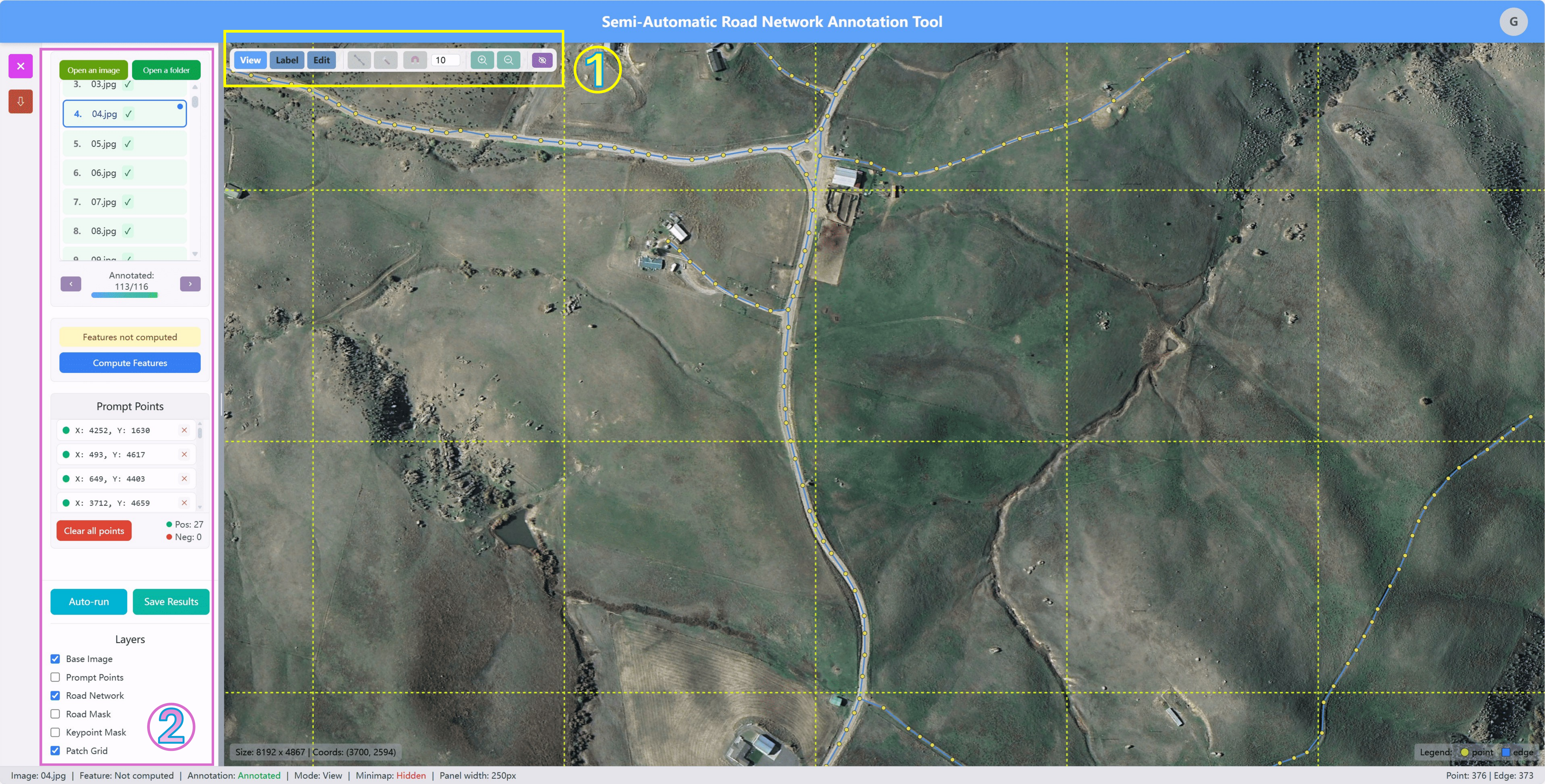}
    \caption{The user interface of our annotation tool, featuring the primary editing toolbar in Region \ding{172} and the main control and information panel in Region \ding{173}, both designed for an intuitive and efficient user experience.}
    \label{fig:appendix_ui}
\end{figure*}

\section{The Interactive Annotation System}
\label{sec:suppl_annotation_system}

To operationalize our interactive framework, we developed a full-featured, web-based annotation tool. This section details its user interface (UI) and the AI-assisted workflow that enables rapid and accurate road network labeling.

\subsection{User Interface Overview}
\label{sec:suppl_ui_overview}

The annotation tool, shown in Figure~\ref{fig:appendix_ui}, is designed for clarity and efficiency. The UI is centered around a main canvas displaying the satellite imagery. The core functionalities are split between two main areas. Region \ding{172} contains the primary editing toolbar, which supports three distinct modes: a \textit{View} mode for free panning and zooming; a \textit{Label} mode, which is the core of the interactive process where users can place positive (left-click) and negative (right-click) prompts; and an \textit{Edit} mode for fine-grained manual adjustments to the generated graph, such as moving vertices or adding and deleting edges. Region \ding{173} serves as the main control and information panel, handling data I/O, providing options for pre-computing image features to accelerate inference, listing prompt point coordinates, managing display layers, and housing the main \enquote{Auto-run} and \enquote{Save Results} buttons. Additionally, a persistent status bar provides helpful auxiliary information, 
such as full image dimensions and real-time cursor coordinates, to aid annotators with spatial awareness and precise editing.

\subsection{Large-Scale Image Inference}
\label{sec:suppl_large_image_inference}

\paragraph{The Challenge and Our Strategy.}
A fundamental challenge is applying our model, which processes 1024$\times$1024 inputs, to high-resolution satellite imagery (e.g., 8K$\times$4K). A naive approach of processing every patch is computationally wasteful, especially in off-road scenes where roads are sparse. To overcome this, we developed a prompt-driven, overlapping patch-based inference pipeline. This strategy ensures that computation is focused exclusively on the user's regions of interest, enabling a smooth and highly efficient interactive experience. The process, detailed in Algorithm~\ref{alg:large_scale_inference}, consists of three key stages.

\paragraph{Stage 1: Prompt-Guided Mask Generation.}
The large input image is first partitioned into a grid of 1024$\times$1024 patches with a significant overlap (e.g., 256 pixels) to prevent artifacts at the boundaries. The inference process is initiated by and centered around the user's sparse prompts. When the user clicks \enquote{Auto-run}, the system first identifies the minimal set of 1024$\times$1024 patches required to cover all prompt points. Inference is run only on this active subset of patches to produce local road and keypoint probability maps. These generated local maps are then seamlessly stitched together into a unified global map for the relevant region. In overlapping areas, pixel values are blended via a weighted average to ensure smooth transitions.

\begin{figure*}[t]
    \centering
    \includegraphics[width=\textwidth]{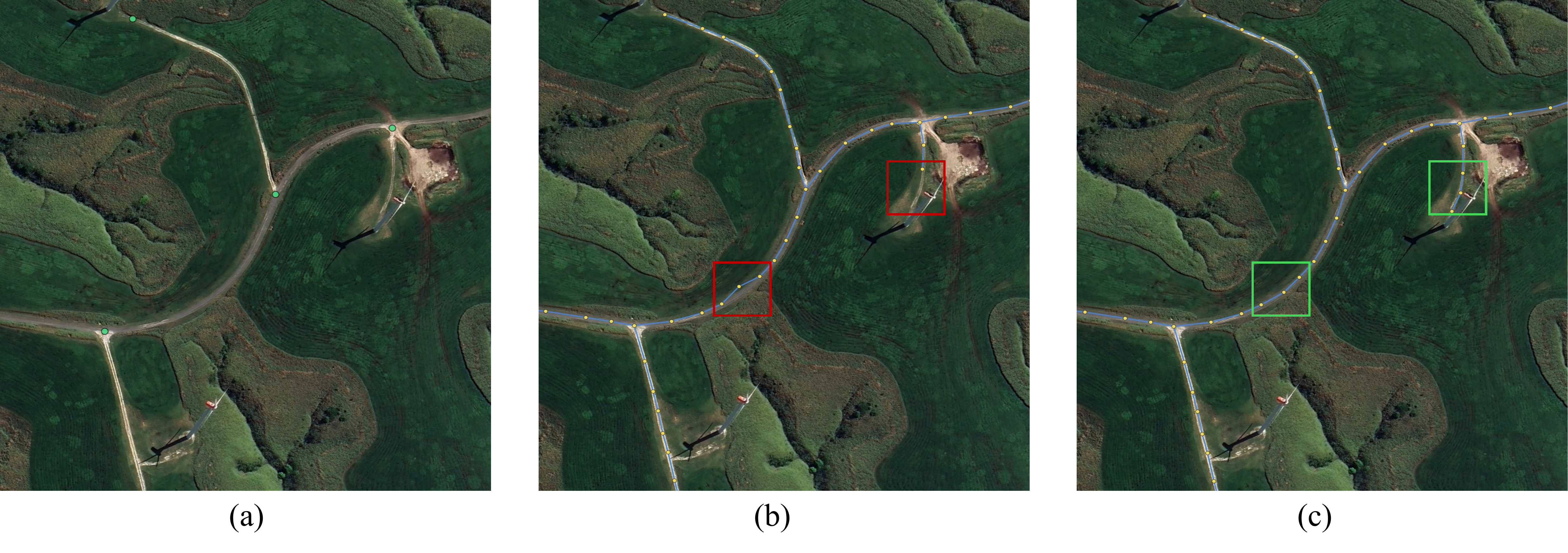}
    \caption{The AI-assisted \enquote{Prompt-Propose-Refine} workflow. (a) An annotator provides sparse prompts at key locations. (b) The model generates a high-quality proposal, which may contain minor errors highlighted by red boxes. (c) The user quickly refines these areas, with the corrected sections shown in green boxes.}
    \label{fig:appendix_workflow}
\end{figure*}

\paragraph{Stage 2: Vertex Extraction and Topological Inference.}
From the fused global keypoint map, a consistent set of candidate vertices is extracted via Non-Maximum Suppression (NMS). With these vertices established, the system then determines their connectivity. To do this efficiently, we revisit each of the previously activated patches. Within a given patch, we consider only the subset of global vertices that fall within its boundaries and use our MaGTopoNet module to predict the probability of edges between them based on the local image context.

\paragraph{Stage 3: Global Edge Aggregation and Final Graph.}
Since patches overlap, a single candidate edge may be evaluated independently in multiple patches, providing a robust ensembling opportunity. In this final stage, we aggregate all connectivity predictions from across the active patches. The scores for each unique edge are averaged, and an edge is included in the final graph only if its average score exceeds a confidence threshold. This ensures the final road network is topologically coherent, resolving local ambiguities and producing a single, unified graph for the user to refine.

\begin{algorithm}[h]
    \caption{Prompt-Driven Large-Scale Inference}
    \label{alg:large_scale_inference}
    \begin{algorithmic}[1]
    \Require Large image $I$, User prompts $P$.
    \vspace{4pt}
    \Ensure Final graph $G=(V, E)$.
    \Statex
    \Function{InferFromPrompts}{$I, P$}
    \State $A_{\text{patches}} \gets \text{IdentifyActivePatches}(P)$
    \State $M_{\text{local}} \gets \{\}$
    \For{patch in $A_{\text{patches}}$}
    \State $M_{\text{local}}[patch] \gets \text{Model.infer}(I, patch)$
    \EndFor
    \State $M_{\text{global}} \gets \text{FuseMasks}(M_{\text{local}})$
    \State $V \gets \text{ExtractVerticesFromMask}(M_{\text{global}})$
    \State $S_{\text{edge}} \gets \{\}$
    \For{patch in $A_{\text{patches}}$}
    \State $V_{\text{patch}} \gets \text{GetVerticesInPatch}(V, patch)$
    \State $S_{\text{patch}} \gets \text{MaGTopoNet}(V_{\text{patch}}, M_{\text{local}}[patch])$
    \State $\text{UpdateEdgeScores}(S_{\text{edge}}, S_{\text{patch}})$
    \EndFor
    \State $E \gets \text{AggregateAndThresholdEdges}(S_{\text{edge}})$
    \State \Return $G=(V, E)$
    \EndFunction
    \end{algorithmic}
\end{algorithm}

\subsection{The ``Prompt-Propose-Refine'' Workflow}

Our system transforms the laborious task of manual road tracing into a highly efficient, human-in-the-loop validation process, which we term the \enquote{Prompt-Propose-Refine} workflow. As illustrated in Figure~\ref{fig:appendix_workflow}, this process unfolds in three intuitive steps.

\begin{itemize}
    \item \textbf{(a) Prompt:} The annotator begins by inspecting the image and placing a sparse set of positive and negative prompts at key topological locations, such as junctions, endpoints, or ambiguous areas.
    \item \textbf{(b) Propose:} After placing the prompts, the annotator clicks \enquote{Auto-run}. The system feeds these prompts to our interactive MaGRoad model, which generates a high-quality road graph proposal in real-time. This proposal serves as a strong baseline, often capturing the majority of the road network correctly.
    \item \textbf{(c) Refine:} The annotator's task is then reduced to curation. They examine the proposal for inaccuracies highlighted by red boxes. By switching to \textit{Edit} mode, they can perform a range of quick corrections, such as repositioning vertices and adding missed connections,  to achieve the final, accurate graph shown in the green boxes.
\end{itemize}
This synergistic \enquote{Prompt-Propose-Refine} paradigm combines the pattern recognition strength of the deep model with the nuanced judgment of a human annotator, achieving both high efficiency and accuracy.

\section{Dataset Organization}
\label{sec:dataset_stats}

\subsection{Data Partitioning Strategy}
\label{subsec:partitioning}

Constructing a high-quality vectorized dataset from large-scale satellite imagery requires a well-designed processing pipeline. Raw gigapixel images (for example, $8k \times 4k$) are too large for direct training, so they are first divided into manageable 1024$\times$1024 patches. To ensure that each patch contains sufficient information and that the overall dataset reflects diverse topological patterns, we adopt a \enquote{Generate–Filter–Select} strategy. This procedure converts raw imagery into a curated collection of samples while reducing repetitive content, as summarized in Algorithm~\ref{alg:partitioning}.
\paragraph{Candidate Generation and Graph Cropping.}
We first employ a sliding window approach to generate a comprehensive pool of candidate patches using two distinct strategies. The primary set A consists of non-overlapping patches sampled on a strict grid with stride 1024 to ensure basic coverage. A supplementary set B is generated using a dense, overlapping sliding window with stride 256 to capture diverse road contexts and shift-variant topologies that might be split across boundaries in the primary grid. A critical challenge here is maintaining graph validity at patch boundaries. We utilize a robust graph cropping algorithm that computes precise geometric intersections between road edges and patch borders. This ensures that all road segments within a patch are properly terminated or connected, preventing invalid topology such as dangling edges or isolated nodes outside the field of view.

\paragraph{Density-Based Filtering.}
In off-road environments, vast regions may contain no road networks. To maintain training efficiency, we filter candidates based on road length density. We compute the total length of road segments within each patch and normalize it by the patch area. Candidates falling below a predefined density threshold $\tau_{\text{density}}$ are identified as empty background and discarded. This step ensures that the model focuses on regions with valid learning signals.

\paragraph{Topology-Aware Diversity Selection.}
A common issue in sliding-window datasets is the inclusion of highly repetitive samples (e.g., identical straight roads shifted by a few pixels). To address this, we introduce a topology-aware selection mechanism using the Weisfeiler-Lehman (WL) Graph Kernel. We first retain all valid patches from the primary Set A. Then, we iteratively evaluate candidates from Set B. For each candidate, we extract its graph topology and compute its WL similarity score against spatially neighboring patches that have already been selected. A candidate is added to the final dataset only if its maximum similarity score is below a threshold $\tau_{\text{sim}}$. This strategy explicitly encourages the inclusion of topologically distinct samples (such as complex junctions or winding paths) while suppressing redundant simple structures.

\begin{algorithm}[t]
    \caption{Topology-Aware Data Partitioning}
    \label{alg:partitioning}
    \begin{algorithmic}[1]
    \Require 
    Large images $\mathcal{I}$, ground-truth graphs $\mathcal{G}$; 
    density threshold $\tau_{\text{density}}$, similarity threshold $\tau_{\text{sim}}$.
    \vspace{4px}
    \Ensure Final dataset $\mathcal{D}$.
    \Statex
    \Function{GenerateAndSelect}{$\mathcal{I}, \mathcal{G}$}
        \State $\mathcal{D} \gets \emptyset$
    
        \For{each $(I, G)$ in $(\mathcal{I}, \mathcal{G})$}
    
            \State \Comment{\textbf{Step 1: Candidate Generation}}
            \State $S_A \gets \textsc{SlidingWindow}(I, G, \text{stride}=1024)$
            \State $S_B \gets \textsc{SlidingWindow}(I, G, \text{stride}=256)$
    
            \State \Comment{\textbf{Step 2: Density Filtering}}
            \State $S_A \gets \{p \in S_A \mid \textsc{Density}(p) \ge \tau_{\text{density}} \}$
            \State $S_B \gets \{p \in S_B \mid \textsc{Density}(p) \ge \tau_{\text{density}} \}$
    
            \State \Comment{\textbf{Step 3: Diversity Selection}}
            \State $\mathcal{D}_{\text{local}} \gets S_A$
            \State \textsc{SortByDensity}( $S_B$ )
    
            \For{each patch $p$ in $S_B$}
                \State $N \gets \textsc{GetSpatialNeighbors}(p, \mathcal{D}_{\text{local}})$
                \State $\text{sim}_{\max} \gets \max_{n \in N} \textsc{WLSim}(p.G, n.G)$

                \If{$\text{sim}_{\max} < \tau_{\text{sim}}$}
                    \State $\mathcal{D}_{\text{local}} \gets 
                        \mathcal{D}_{\text{local}} \cup \{p\}$
                \EndIf
            \EndFor
    
            \State $\mathcal{D} \gets \mathcal{D} \cup \mathcal{D}_{\text{local}}$
        \EndFor
    
        \State \Return $\mathcal{D}$
    \EndFunction
    
    \end{algorithmic}
\end{algorithm}

\subsection{Dataset Statistics}
After applying our partitioning and selection strategy, the final WildRoad dataset comprises a total of 9,274 curated patches. Table~\ref{tab:dataset_stats} details the distribution across the training, validation, and test sets. The training set contains 6,448 patches with over 4,000 km of road network and more than 11,000 intersections, providing a rich source of topological variety for model learning. The validation and test sets are similarly structured, ensuring a robust evaluation of generalization capability across diverse off-road scenarios.

\begin{table}[h]
    \centering
    \caption{Dataset statistics for road network analysis.}
    \label{tab:dataset_stats}
    \begin{tabular}{c|cccc}
    \toprule
    Dataset & Files & Length (km) & Intersections & Endpoints \\
    \midrule
    Train   & 6,448 & 4,104.99 & 11,172 & 35,530 \\
    Val     & 1,493 &   951.61 &  2,573 &  8,298 \\
    Test    & 1,333 &   810.95 &  2,180 &  6,941 \\
    \bottomrule
    \end{tabular}
\end{table}


\end{document}